\documentclass[lettersize,journal]{IEEEtran}
\usepackage{amsmath,amsfonts}
\usepackage{algorithmic}
\usepackage{array}
\usepackage[caption=false,font=normalsize,labelfont=sf,textfont=sf]{subfig}
\usepackage{textcomp}
\usepackage{multirow}
\usepackage{float}
\usepackage{stfloats}
\usepackage{hyperref}
\hypersetup{
    colorlinks=true,
    linkcolor=blue,
    citecolor=blue,
    urlcolor=blue,
    }
\usepackage{makecell}
\usepackage{placeins}
\usepackage{url}
\usepackage{graphicx}
\def\BibTeX{{\rm B\kern-.05em{\sc i\kern-.025em b}\kern-.08em
    T\kern-.1667em\lower.7ex\hbox{E}\kern-.125emX}}
\usepackage{balance}

\begin{document}
\title{AI-Enhanced Cognitive Behavioral Therapy: Deep Learning and Large Language Models for Extracting Cognitive Pathways from Social Media Texts}
\author{Meng Jiang, Yi Jing Yu, Qing Zhao, Jianqiang Li, Changwei Song, Hongzhi Qi, Wei Zhai, Dan Luo, Xiaoqin Wang, Guanghui Fu, Bing Xiang Yang\textsuperscript{*}
\thanks{This work was supported by grants from the National Natural Science Foundation of China (grant numbers:72174152, 72304212 and 82071546), Fundamental Research Funds for the Central Universities (grant numbers: 2042022kf1218; 2042022kf1037), the Young Top-notch Talent Cultivation Program of Hubei Province.}

\thanks{Meng Jiang,  Qing Zhao, Jianqiang Li, Changwei Song, Hongzhi Qi, Wei Zhai are with School of Software Engineering, Beijing University of Technology, Beijing, China.}
\thanks{Yi Jing Yu, Dan Luo, Xiaoqin Wang, Bing Xiang Yang are with School of Nursing, Wuhan University, Wuhan, China.}
\thanks{Guanghui Fu is with Sorbonne Universit\'{e}, Institut du Cerveau – Paris Brain Institute - ICM, CNRS, Inria, Inserm, AP-HP, H\^{o}pital de la Piti\'{e}-Salp\^{e}tri\`{e}re, F-75013, Paris, France.}
\thanks{Guanghui Fu is supported by a Chinese Government Scholarship provided by the China Scholarship Council (CSC).}
\thanks{\textsuperscript{*}Corresponding author: Bing Xiang Yang (\url{yangbx@whu.edu.cn})}
}

\markboth{Preprint}%
{Jiang \MakeLowercase{\textit{et al.}}: }

\maketitle

\begin{abstract}

Cognitive Behavioral Therapy (CBT) is an effective technique for addressing the irrational thoughts stemming from mental illnesses, but it necessitates precise identification of cognitive pathways to be successfully implemented in patient care. In current society, individuals frequently express negative emotions on social media on specific topics, often exhibiting cognitive distortions, including suicidal behaviors in extreme cases. Yet, there is a notable absence of methodologies for analyzing cognitive pathways that could aid psychotherapists in conducting effective interventions online. In this study, we gathered data from social media and established the task of extracting cognitive pathways, annotating the data based on a cognitive theoretical framework. We initially categorized the task of extracting cognitive pathways as a hierarchical text classification with four main categories and nineteen subcategories. Following this, we structured a text summarization task to help psychotherapists quickly grasp the essential information. Our experiments evaluate the performance of deep learning and large language models (LLMs) on these tasks. The results demonstrate that our deep learning method achieved a micro-F1 score of 62.34\% in the hierarchical text classification task. Meanwhile, in the text summarization task, GPT-4 attained a Rouge-1 score of 54.92 and a Rouge-2 score of 30.86, surpassing the experimental deep learning model's performance. However, it may suffer from an issue of hallucination. We have made all models and codes publicly available to support further research in this field: \url{https://github.com/JiangMeng-JM/Cognitive-Pathways---Deep-Learning}.

\end{abstract}


\section{Introduction} \label{sec:introduction}

\IEEEPARstart{D}{epression}, being a prevalent mental health disorder, constitutes one of the foremost causes of mental-health-related disabilities worldwide. According to the World Health Organization (WHO) statistics, an estimated 3.8\% of the global population grapples with depression~\cite{WHO}. The situation is not optimistic in China either, as the country alone accounts for a significant 6.9\% prevalence of such cases~\cite{huang2019prevalence}, this also underscores the burgeoning mental health challenges. Consequently, monitoring and intervening in depressive emotions have become increasingly crucial. In contemporary society, people frequently share their emotions and experiences on social media platforms such as Sina Weibo, X (Twitter), and Reddit. The anonymity feature provided by online forums makes individuals with negative emotions more willing to disclose their personal feelings and experiences. In the context of themes such as ``Depression'' and ``Suicide'', where individuals are more likely to express negative emotions, demonstrate cognitive distortions through their language, and in some cases, reveal suicidal thoughts~\cite{robinson2016social}. Cognitive Behavioral Therapy (CBT) is recommended by clinical guidelines as a key treatment for depression due to its effectiveness in correcting cognitive distortions, thereby reducing negative emotions~\cite{cuijpers2023cognitive}. A pre-requirement for conducting CBT involves identifying cognitive pathways in the language of those with cognitive distortions. According to Ellis's Rational-Emotive Behavior Therapy (REBT)~\cite{ellis2007practice}, these can be categorized into four primary nodes: \textbf{A}ctivating event, \textbf{B}elief (cognitive distortions), \textbf{C}onsequence, and \textbf{D}isputation, abbreviated as ABCD. However, inexperienced psychotherapists often face challenges in accurately identifying these nodes, leading to significant obstacles for precise subsequent interventions~\cite{stein2022psychiatric}. This highlights a critical need for effective tools within the field.

Deep learning, trained on extensive datasets, has significantly surpassed traditional algorithms in performance~\cite{young2018recent}, particularly showing remarkable potential in Natural Language Processing (NLP) tasks in recent years~\cite{otter2020survey}. In the field of psychology, deep learning models have been employed to analyze people's language like sentiment analysis~\cite{alharbi2019twitter, nandwani2021review} and suicide risk identification~\cite{tadesse2019detection, fu2021distant}. Given the challenges of collecting high-quality data in this domain, the development of pre-trained models and transfer learning techniques has been crucial~\cite{singh2020realising, aragon2023disorbert, zhai2024chinese}. The development of large language models (LLMs) techniques has garnered increasing interest and application in psychology~\cite{he2023towards}. Qi et al.~\cite{qi2023evaluating} conduct experiments to compare LLMs and supervised learning in cognitive distortion identification and suicide risk classification. The experimental results indicate that LLMs struggle with accurately identifying complex cognitive distortions in Chinese social media data, suggesting that deep learning algorithms remain the preferred solution for complex psychological tasks. To the best of our knowledge, specific studies and methods focused on cognitive pathways recognition are currently lacking.

\begin{figure*}[!t]
\centering
\includegraphics[width=4in]{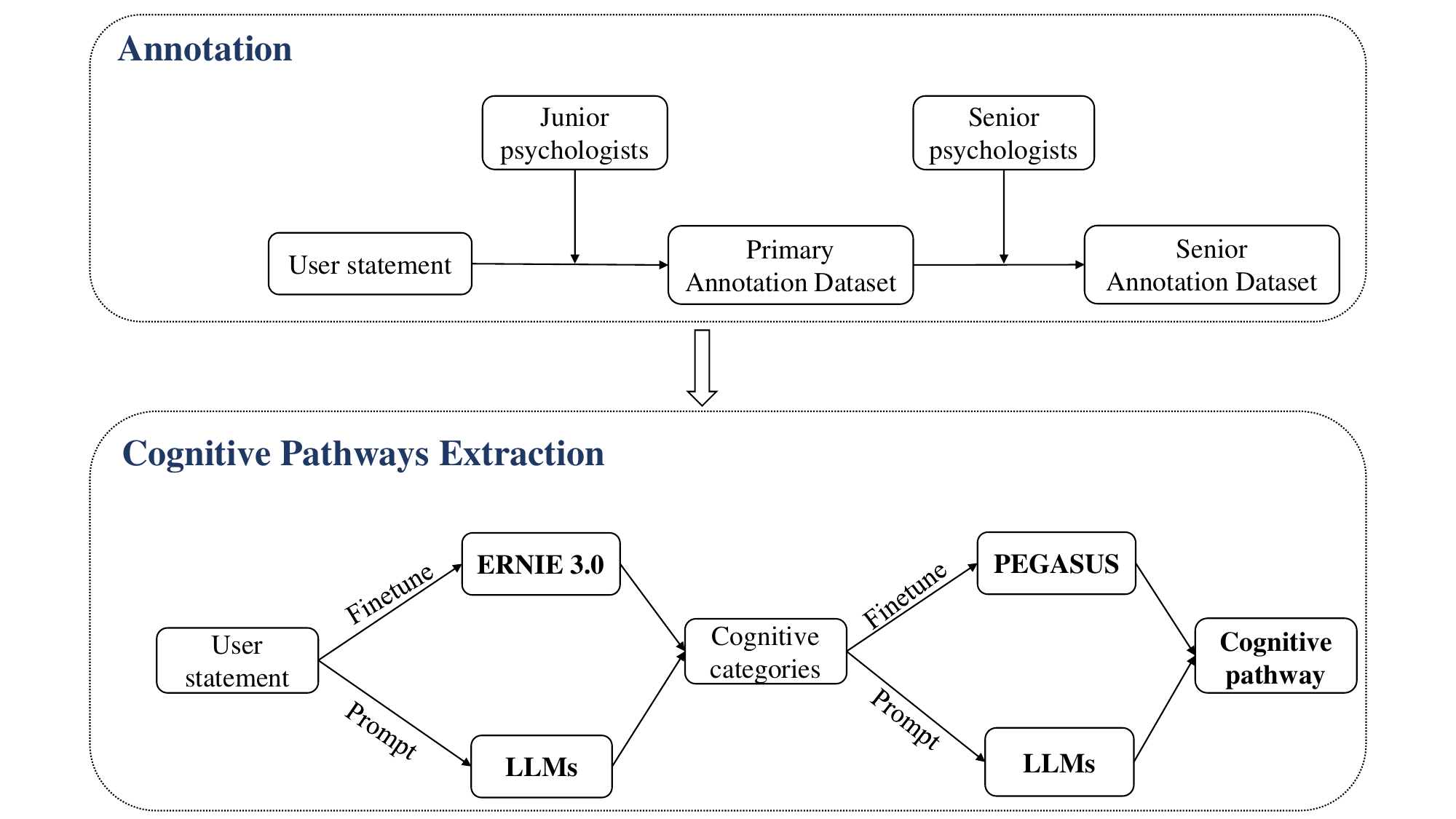}
\caption{This figure outlines the workflow of the research. Initially, experts at various experience levels collaborate to annotate the data. Subsequently, different methods (deep learning model or LLMs) are developed to classify the cognitive categories from the user statement. In the final step, the results of these pathways are summarized to produce a final cognitive pathway to help psychotherapists understand quickly.}
\label{fig:overall_flow}
\end{figure*}

As shown in Figure~\ref{fig:overall_flow}, in this study, we collect data from social media and annotate it to develop models for extracting cognitive pathways. As previously discussed, cognitive pathways can be conceptualized into four parent nodes (ABCD) and various sub-nodes, based on psychological cognitive theories~\cite{ellis2007practice}. Therefore, we frame the first task as a hierarchical text classification, categorizing text into four parent and nineteen child nodes. Given the frequent redundancy and lack of clarity in social media statement, we introduce the second task: text summarization, aimed at providing psychotherapists with quick and clear access to the cognitive pathways framework. We explored these challenges using two approaches: pre-trained language models (ERNIE 3.0~\cite{ernie3_sun2021ernie} and PEGASUS~\cite{pegasus_zhang2020pegasus}) and the GPT series of LLMs~\cite{openai2023gpt4}. The results indicate that deep learning models excel in extracting cognitive pathways, whereas GPT-4 outperforms in text summarization. We have made the trained models publicly available to support further analysis by healthcare professionals, researchers, and other interested individuals.

\section{Background knowledge} \label{sec:background} 
According to Beck's cognitive theory, cognitive factors are crucial in the development and maintenance of depression~\cite{beck1989cognitive}. CBT is recognized as an effective psychotherapeutic approach that boosts emotional and psychological health by addressing and modifying negative thought patterns and behaviors. The fundamental concept of CBT suggests that altering irrational thoughts can positively affect an individual's mood and actions. In this therapy, the psychotherapist collaborates with the patient to pinpoint cognitive distortions, applying cognitive reframing and behavior modification strategies for improvement. Research has validated CBT's substantial effectiveness in treating various psychological disorders, including depression and anxiety~\cite{hofmann2012efficacy, tolin2010cognitive}.

\begin{figure*}[!t]
\centering
\includegraphics[width=0.85\linewidth]
{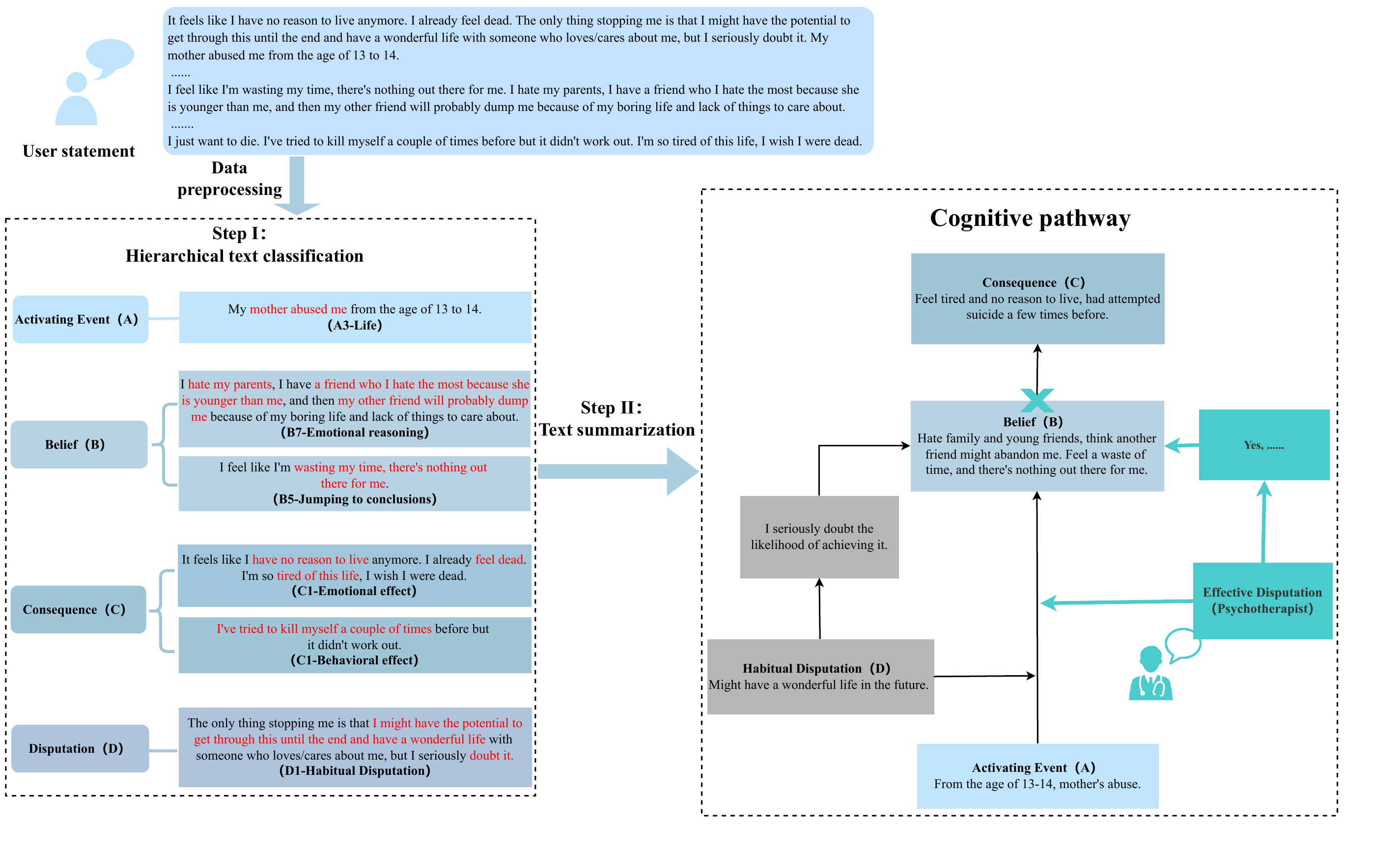}
\caption{The figure presents an English translation of a user's statement from Chinese social media, highlighting suicidal ideation and cognitive distortions. It demonstrates how the user's statement is segmented into ABCD components in CBT. By analyzing statements, psychotherapist can effectively counter irrational belief (B) and prevent habitual disputation (D) through effective disputation (D), thereby gradually amending cognitive distortions. Note that due to space constraints, some irrelevant information in the user statement is not shown in the translation.}
\label{fig:cognitive_pathways_example}
\end{figure*}

The ABCD model stands as the cornerstone framework within CBT for analyzing and addressing cognitive processes, as outlined by Sarracino et al.~\cite{sarracino2017rebt} and Selva et al.~\cite{selva2021albert}. This model posits that the emotional and behavioral reactions (C: Consequence) are not directly caused by events (A: Activating event) but by their beliefs and interpretations (B: Belief) of these events. Through the therapeutic process of challenging and disputing (D: Disputation) irrational beliefs, the psychotherapist aids the patient in developing more effective coping mechanisms, thereby rendering CBT a more tangible and targeted approach. Accurate and effective identification of the ABCD framework is crucial for the success of CBT. This model is defined in details as follows, and an example can be seen in Figure~\ref{fig:cognitive_pathways_example}.

\begin{itemize}
    \item \textbf{(A) Activating event:} An activating event is an event or situation that the patient is facing, it could be a conversation, a work assignment, a family argument or an accident. For instance, Figure~\ref{fig:cognitive_pathways_example}, step~I~(A), includes the statement, ``From the time I was 13-14 years old, my mother abused me''. 
    \item \textbf{(B) Belief:} When confronted with an activating event, individuals form beliefs which can be either rational or irrational, influenced by subjective experiences and thought patterns. In individuals with depression, irrational beliefs often emerge. An example is illustrated in step~I~(B) of Figure~\ref{fig:cognitive_pathways_example}: ``I feel like I'm wasting my time; there's nothing out there for me''.
    Burns et al.~\cite{burns1999feeling} summarized irrational beliefs common in people with depression, namely cognitive distortions.
    \item \textbf{(C) Consequence:} The third component involves the consequences of irrational beliefs, which can be emotional, behavioral, or both. For example, Figure~\ref{fig:cognitive_pathways_example} step~I~(C) demonstrates that the user experiences both emotional consequences ``It feels like I have no reason to live anymore... tired of this life, I wish I were dead.'' and behavioral consequences ``I've tried to kill myself a couple of times before...''.
    \item \textbf{(D) Disputation:} Disputation come in two forms: habitual and effective. Habitual disputation happens automatically, often without conscious thought, whereas effective disputation involves deliberate rebuttal through reasoned thinking and evidence presentation, which can lead to more constructive outcomes. Individuals with depression commonly exhibit habitual disputation, as illustrated in the step~I~(D) in Figure~\ref{fig:cognitive_pathways_example}. This user expresses habitual disputation with the statement ``The only thing stopping me is... but I seriously doubt it.'', necessitating psychotherapeutic intervention to foster effective disputation.
\end{itemize}

\section{Related work} \label{sec:related} 

\subsection{AI-assisted cognitive behavioral therapy} \label{sec:related:ai_cognitive} 
Cognitive Behavioral Therapy (CBT) is a problem-focused, empirically based short-term psychotherapy that alleviates and treats various psychological issues by assisting patients in altering maladaptive thought patterns and behavioral habits~\cite{payne2010increasing}. It stands as one of the most extensively utilized and efficacious psychological interventions for depression~\cite{foreman2016cognitive}. 
With the advancement of computer technology and mobile healthcare, computerized cognitive behavioral therapy (CCBT) has emerged as an alternative mode of delivery, attracting the attention and in-depth research~\cite{bailey2018theoretical, heng2021rewind}. This innovative CCBT tool compensates to some extent for the limitations of traditional CBT, which requires face-to-face implementation~\cite{liu2021efficacy}. However, despite the advantages and convenience of CCBT in providing psychological support, it also faces several challenges, including a lack of medical diagnostic capability, insufficient levels of personalization, and poor interactivity. The development of AI, particularly LLMs like the series of GPT~\cite{brown2020language, openai2023gpt4}, has opened up new avenues for tackling these challenges. By analyzing patients' language, behavior, and other relevant data, AI can offer diagnostic support to clinical professionals, helping them diagnose mental health issues more quickly and accurately, and assess the severity of symptoms~\cite{graham2019artificial}. Therefore, many studies have focused on integrating AI with CBT to better understand and treat mental health issues. For instance, Vuyyuru et al.~\cite{vuyyuru2023transformer} introduced a novel Trans-CNN hybrid model that effectively integrates Transformer and CNN architectures, demonstrating superior performance in recognizing psychological disorders compared to traditional CBT methods and independent Transformer or CNN models, thus elevating the efficacy of CBT in psychological assessment and intervention. Furthermore, in enhancing interactivity, numerous researchers are devoted to developing various tools, such as chatbots and conversational agents, to emulate the communicative style of human psychotherapists, thereby offering efficient, personalized, and readily accessible CBT treatments~\cite{fitzpatrick2017delivering, rani2023mental}. Currently, cognitive-behavioral interventions heavily rely on psychotherapists' continuous assessment of patients, imposing significant demands on psychotherapists. However, there exists a pronounced deficiency in mental health professionals, particularly in terms of high-caliber personnel. To address this issue, Fu et al.~\cite{fu2023enhancing} introduced a psychological counseling support system based on a LLM, aiming to assist novice counselors and volunteers in providing online psychological support, thereby alleviating some of the pressure caused by the current resource constraints. Sharma et al.~\cite{sharma2023cognitive} focused on cognitive reframing, a core intervention in CBT. They utilize the language models (LMs) to design a system that assists individuals in restructuring negative thinking, thereby facilitating self-guided psychological interventions. A randomized field study conducted on the Mental Health America website confirms the effectiveness of the system. Sharma et al.~\cite{sharma2023facilitating} emphasized the use of LMs to support individuals in completing each step of cognitive restructuring (such as identifying thinking traps and writing reframed thoughts). Furthermore, it assessed the fairness of the system across different demographic groups, highlighting the importance of personalization, interactivity, fairness, and safety in designing self-guided mental health interventions. While the aforementioned studies showcase the vast potential of AI technology in the CBT field, existing cognitive behavioral interventions typically only address overall cognitive issues in the population with depression. There is a dearth of understanding of individual-specific, unique cognitive distortions, and cognitive pathways, making precise intervention difficult. As underscored by Huibers et al.~\cite{huibers2021road}, it is imperative to develop personalized psychotherapy tailored to the specific needs of each individual patient. Therefore, the effective utilization of AI to explore personalized cognitive features and pathways associated with depression is imperative. 

\subsection{Hierarchical text classification} \label{sec:related:classify}
Text classification is a classic task in NLP involving the categorization of given text into predefined classes or labels~\cite{minaee2021deep}. However, in real-world scenarios like news categorization, patent classification, and the categorization of academic reports, there is a need to organize text into a set of labels that are structured hierarchically~\cite{silla2011survey}. Currently, existing methodologies for the HTC task are primarily categorized into two approaches: local and global~\cite{zhou2020hierarchy}. Early research on HTC primarily focused on the local approach, which treats each category as an independent sub-category classification task without considering the overall hierarchical structure. Additionally, some research has concentrated on the overall hierarchical structure and attempted to capture the relationships between categories to enhance classification accuracy. For instance, Peng et al.~\cite{peng2018large} proposed a Deep Graph-CNN based on deep learning model and regularize the deep architecture with the dependency among labels in order to further leverage the hierarchy of labels. Mao et al.~\cite{mao2019hierarchical} proposed a global framework called HiLAP, which incorporates a novel end-to-end reinforcement learning technique for better consideration of label hierarchy during both training and inference processes. This framework shows more flexibility and generalization capacity than previous global approaches, as it imposes no constraints on the hierarchy structure or object labels assigned. Additionally, some studies have combined local and global approaches~\cite{wehrmann2018hierarchical}. However, the aforementioned method requires a large amount of carefully annotated training data to achieve satisfactory performance, which will result in excessively high costs. There is an urgent need to construct models that can achieve stronger adaptability with less training data. Therefore, Meng et al.~\cite{meng2019weakly} proposed a neural approach called WeSHClass for weakly supervised hierarchical text classification. It only requires providing some weak supervision signals, through a pseudo document generation module, to generate high-quality pseudo-training data, significantly alleviating the training data bottleneck issue. In recent years, with the emergence of pre-trained language models, they have been capable of leveraging unsupervised learning methods to pre-train on massive datasets, thereby acquiring rich linguistic knowledge and representation capabilities. Subsequently, by fine-tuning the models with relatively small amounts of domain-specific data, they can be transferred to various downstream tasks. This strategy greatly alleviates the challenge of scarce training data~\cite{han2021pre}. Chiorrini et al.~\cite{chiorrini2021emotion} demonstrated that the capabilities of BERT that can achieve good performance on text classification. 
In our study, we also faced the challenge of insufficient data; consequently, we employed a pre-trained model for the HTC task to extract cognitive pathways from social media statement.

\subsection{Automatic text summarization} \label{sec:related:summary}
Text summarization aimed at alleviating information overload by crafting concise yet informative summaries from input text. It can be broadly classified into two types: extractive and abstractive summarization. The extractive summarization method generate summaries by selecting key concepts and topics from the original text and assembling them to create a concise summary, while the abstractive summarization method generate new expressions by understanding and interpreting the original text, resulting in summaries that are closer to human-generated summaries~\cite{chopra2016abstractive}. Many deep neural network models have been proposed for abstractive summarization. Rush et al.~\cite{rush2015neural} proposes a data-driven approach for sentence summarization, which uses a local attention-based model to generate each word of the summary given the input sentence. Chopra et al.~\cite{chopra2016abstractive} uses a conditional RNN to generate the summary, this is an extended study of Rush et al.~\cite{rush2015neural}. 
They employed a seq2seq architecture featuring a CNN in the encoder and an RNN in the decoder for the sentence-level abstractive summarization task. 
Subsequently, the introduction of the Transformer architecture~\cite{transformer_vaswani2017attention} led researchers to discover that Transformer models, when pre-trained on large text corpora through self-supervised learning, can achieve enhanced performance in tasks related to natural language understanding and text generation~\cite{radford2018improving}. Consequently, researchers have pre-trained many models based on the transformer architecture and applied them to the field of text summarization, including models such as T5~\cite{raffel2020exploring}, Bart~\cite{lewis2020bart}, PEGASUS~\cite{pegasus_zhang2020pegasus}. Zhang et al.~\cite{pegasus_zhang2020pegasus} introduced the first pre-trained model PEGASUS specifically designed for abstractive summarization, unlike BART~\cite{lewis2020bart} and T5~\cite{raffel2020exploring}, the PEGASUS model masks multiple complete sentences based on importance rather than smaller contiguous text segments during pre-training, which achieved excellent results with just 1,000 fine-tuning examples. With the advancement of LLMs, people have also validated their performance in text summarization tasks. Goyal et al.~\cite{goyal2022news} compared prompt-based GPT-3 with fine-tuned models in the news summarization task and demonstrated that news summaries generated by GPT-3 are more favorably received by humans. Pu et al.~\cite{Pu2023summarization} found that summaries generated by LLMs exhibited remarkable fluency, authenticity, and adaptability compared to human-written reference summaries and those generated by fine-tuned models in various summarization tasks, especially in specialized and less common summarization scenarios.
However, LLMs have the potential to generate hallucinatory content, which compromises their reliability in practical applications~\cite{xu2024hallucination, tonmoy2024comprehensive}. Consequently, the choice between deep learning methods and LLM-based approaches remains a subject of debate.

\section{Methods} \label{sec:methods}

\subsection{Task definition} \label{sec:taskDefinition}

\paragraph{Hierarchical text classification} Given a user statement text sequence $X$ containing $n$ sentences, $X = \{s_1, s_2, ..., s_n\}$, our goal is to extract the cognitive pathways of this user through two tasks: hierarchical text classification and text summarization. First, based on the hierarchical nature of the ABCD model, we define the categories into sets of parent and child categories. The set of parent categories is denoted as: $P = \{p_a, p_b, p_c, p_d\}$ for ABCD model. Then, for each parent category $p_i$, we define its corresponding set of child categories as $C_{p_i} = \{c_1^{{p_i}}, c_2^{{p_i}}, ..., c_{k_{p_i}}^{p_i}\}$, where $k_{p_i}$ is the number of the child categories under parent category $p_i$. The label of a sentence $s_n$ is represented as $L_{s_n}=\{(p_a, c_1^{p_a}, ..., c_{k_{p_a}}^{p_a}), \cdots, (p_d, c_1^{p_d}, ..., c_{k_{p_d}}^{p_d})\}$. All categories, along with their hierarchical structure and distribution, are detailed in Table~\ref{tab:dataset}.

\paragraph{Text summarization} After applying the hierarchical classification model to each sentence $s_i$ in the user statement $X$, we categorize each sentence into a parent category represented. For each parent category $p_i$, we aggregate sentences classified under this category into a new, composite sentence $S_{p_i}$, and it is formed by concatenating sentences: $S_{pi}=\{s_1^{p_i} s_2^{p_i} ... s_m^{p_i}\}$. The objective is to generate an abstractive summary for the sentences grouped under each parent category. The summary for category $p_i$ is represented as $A_{p_i}=TextSummarizer(S_{p_i})$, where $S_{pi}$ is the input composite sentence. 

After the two step operation of hierarchical text classification and abstractive summarization, the final output of cognitive pathways will be a set of summaries $\{A_a, A_b, A_c, A_d\}$.

\subsection{ERNIE 3.0 for hierarchical text classification}
In the field of psychology, tasks often suffer from a lack of training data, making it challenging to train deep learning models from scratch. A common solution to this issue is to finetune from a pre-trained model. In our research, we opted to fine-tune the ERNIE 3.0 (Enhanced Representation through Knowledge Integration) model~\cite{ernie3_sun2021ernie}, for our task of extracting cognitive pathways. ERNIE 3.0 serves as the foundation for the LLM as ERNIE Bot\footnote{Baidu ERNIE bot (Wenxin Yiyan): \url{https://yiyan. baidu.com/welcome}}. ERNIE 3.0 combines an auto-regressive network with an auto-encoding network, training on a 4TB corpus that includes both plain texts and a comprehensive knowledge graph. This model introduces a novel pre-training framework known as Continual Multi-Paradigms Unified Pre-training. This framework comprises two core modules: the universal representation module and the task-specific representation module. The universal representation module utilizes a multi-layer Transformer-XL~\cite{dai2019transformer}, serving as its backbone network. Unlike the standard Transformer model~\cite{transformer_vaswani2017attention}, Transformer-XL is designed to support longer sequence lengths, enhancing its ability to process and understand extended textual contexts. Similarly, the task-specific representation module employs a multi-layer Transformer-XL structure. Its design focuses on capturing the high-level semantic representations for various task paradigms.

In the fine-tuning phase, we froze the universal representation module and proceeded to fine-tune the task-specific representation module. We then linked it with a fully connected layer that acts as a classifier for the hierarchical text classification task.

\subsection{PEGASUS for text summarization} \label{sec:pegasus}

We fine-tuned the PEGASUS model~\cite{pegasus_zhang2020pegasus} for our text summarization task. PEGASUS builds on the Transformer model~\cite{transformer_vaswani2017attention}, employing the encoder-decoder structure. Within this framework, the encoder processes and understands the input text, whereas the decoder is responsible for producing summaries.

During the pretraining process, ``important sentences'' from the input document are masked. The model is then tasked with using the remaining sentences to generate the omitted sentences, attempting to reconstruct the original document. The pretraining input comprises documents with these sentences removed, while the output is the concatenation of the missing sentences. However, such a challenging task encourages the model to learn the facts present in the corpus and how to distil information from the whole document in order to generate outputs that are very similar to the fine-tuned summary task. Consequently, PEGASUS demonstrates outstanding performance on small-scale datasets, achieving effective fine-tuning with limited data.

\subsection{LLM prompt}\label{sec:prompt}

LLMs have shown exceptional performance across diverse domains. Yet, their effectiveness in extracting cognitive pathways in psychology is not well-established. We used a structured prompt designed to guide LLMs to extract cognitive pathways from user statements.
The prompt design incorporates several elements: defining the role, specifying the task, outlining the structure of cognitive pathways, and defining the output format. The task definition includes two steps: extracting cognitive pathways from the user's statement and then summarizing into ABCD structure based on the identified pathways. We also compared performance with supervised learning methods.

\section{Experiments} \label{sec:experiments}

\subsection{Datasets} \label{sec:experiments:datasets}
Our dataset comprises data from three sources: comments from the ``ZouFan treehole'' on Weibo\footnote{\url{https://www.weibo.com/xiaofan116?is_all=1}}, and posts from the ``SuicideWatch'' and ``depression'' subreddits on Reddit\footnote{\url{https://www.kaggle.com/datasets/nikhileswarkomati/suicide-watch}}. We translated the Reddit-derived data into Chinese by google translate. During preprocessing, we excluded irrelevant sentences, such as those with non-ASCII characters or posts shorter than 100 words. To construct the dataset, 6 annotators, including 4 junior and 2 senior psychologists, annotated 555 posts.

To ensure the quality of our labeling, we implemented two key preparatory processes. The flowchart is shown in the ``Annotation'' part of Figure~\ref{fig:overall_flow}.
\begin{itemize}
    \item Development of the annotation guide: We began by identifying essential elements and drafting an initial guide. A consultation meeting with experts and annotators then refined this guide, discussing its structure, theoretical basis, and core concepts with examples. Through iterative feedback and revisions, we finalized the cognitive feature labels and the annotation guide.
    \item Labeling consistency training: We conducted training sessions for medical students with mental health expertise, led by a psychotherapist, to understand the labeling process and maintain consistency. Annotators worked in pairs according to the guide, with any discrepancies resolved through expert judgment until consistency standards were achieved.
\end{itemize}

For the cognitive pathways extraction tasks, we divided the dataset into training, validation, and test sets in a 6:2:2 ratio, resulting in 333, 111, and 111 posts for each set, respectively.
Subsequently, convert the annotated dataset into formats suitable for hierarchical text classification models and text summarization models. In total, the dataset for hierarchical text classification comprises 4742 sentences, distributed across the training, validation, and test sets with 2835, 932, and 975 sentences, respectively. The detailed data distribution is presented in Table~\ref{tab:dataset}.

\begin{figure}[!t]
\centering
\includegraphics[width=3.5in]{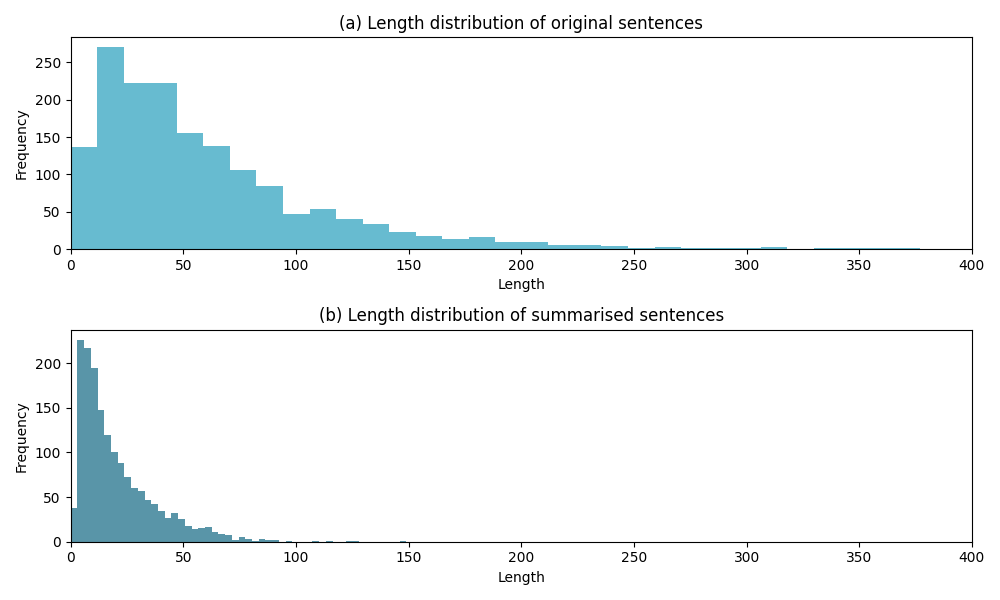}
\caption{Distribution of sentence lengths in the dataset for original sentence lengths (a) and human-written reference summaries (b).}
\label{fig:task_summary_length}
\end{figure}

The dataset for text summarization contains 1643 sentence-summary pairs, divided into training, validation, and test sets with 983, 326, and 334 pairs, respectively. A comparison of sentence lengths before and after summarization is illustrated in Figure~\ref{fig:task_summary_length}.

\begin{table*}
\centering
\caption{Distribution of experimental data for hierarchical text classification task in sentence level. The parent nodes representing the cognitive pathways framework ABCD, and the child nodes corresponding to its sub-categories.}
\scalebox{0.8}{
\begin{tabular}{|l|l|c|c|c|c|c|}
\hline
\multicolumn{1}{|l|}{Parent nodes}& Child nodes                       & Train set & Val set & Test set & \multicolumn{2}{c|}{Sum} \\
\hline
\multirow{5}{*}{(A) Activating Event} & Disease symptom                & 196       & 66      & 69       & 331 & \multirow{5}{*}{1590}   \\
                                      & Social relation                & 200       & 69      & 70       & 339 &                         \\
                                      & Life                           & 214       & 72      & 68       & 354 &                         \\
                                      & Study and work                 & 184       & 69      & 60       & 313 &                         \\
                                      & Emotional                      & 154       & 47      & 52       & 253 &                         \\ 
\hline
\multirow{10}{*}{(B) Belief}          & All-or-nothing thinking            & 51        & 14      & 16       & 81  & \multirow{10}{*}{1803}  \\
                                      & Over-generalization             & 101       & 39      & 35       & 175 &                         \\
                                      & Mental filter               & 54        & 9       & 16       & 79  &                         \\
                                      & Disqualifying the positive     & 120       & 46      & 39       & 205 &                         \\
                                      & Jumping to conclusions         & 230       & 75      & 77       & 382 &                         \\
                                      & Magnification and minimization & 55        & 14      & 20       & 89  &                         \\
                                      & Emotional reasoning            & 206       & 64      & 74       & 344 &                         \\
                                      & Should statements              & 19        & 6       & 5        & 30  &                         \\
                                      & Labeling and mislabeling                       & 171       & 59      & 64       & 294 &                         \\
                                      & Blaming oneself/others         & 80        & 20      & 24       & 124 &                         \\ 
\hline
\multirow{2}{*}{(C) Consequence}     & Emotional effect               & 361       & 118     & 123      & 602 & \multirow{2}{*}{1071}   \\
                                      & Behavioral effect              & 276       & 94      & 99       & 469 &                         \\ 
\hline
\multirow{2}{*}{(D) Disputation}      & Habitual disputation           & 89        & 28      & 33       & 150 & \multirow{2}{*}{278}    \\
                                      & Effective disputation          & 74        & 23      & 31       & 128 &                         \\
\hline
\end{tabular}
}
\label{tab:dataset}
\end{table*}

\subsection{Evaluation metrics} \label{sec:evaluation_metrics}
For the hierarchical text classification task, we use precision, recall, and the F1-score as evaluation metrics. Given the issue of category imbalance in our dataset, we utilize micro-averaging to compute these metrics, which considers the aggregate contribution of all classes to compute the average metric.

In the task of text summarization, we utilize three variants of the ROUGE metric for evaluation: ROUGE-1, which assesses the overlap of unigrams between the generated text and the reference text; ROUGE-2, which measures the overlap of bigrams; and ROUGE-L, which evaluates the longest common subsequence. Additionally, we use the BLEU-4 metric, which directly quantifies the 4-gram overlap between the generated text and the reference summary, to evaluate the similarity between them.

\subsection{Implementation details} \label{sec:Implementation details}

We are using the PaddleNLP, a natural language processing (NLP) toolkit\footnote{\url{https://github.com/PaddlePaddle/PaddleNLP}}, which comes with built-in weight parameters and corresponding model architectures for ERNIE~\cite{ernie3_sun2021ernie} and PEGASUS~\cite{pegasus_zhang2020pegasus} pre-trained models.
For fine-tuning the ERNIE 3.0~\cite{ernie3_sun2021ernie} for hierarchical text classification, we set the learning rate to 3e-5, batch size to 32, and fine-tuning 40 epochs. 
To avoid instability from high learning rates, we employ a warmup strategy with warmup steps set to 150, improving model convergence. Early stopping is applied after 10 epochs without improvement to prevent overfitting.

For text summarization, we employed the Pegasus~\cite{pegasus_zhang2020pegasus} for fine-tuning. We utilized ReLU as activation function, set a learning rate of 5e-5 and a batch size of 32, alongside a 0.02 learning rate warm-up to speed up convergence. We use the Adam optimizer, with an epsilon of 1e-6 and weight decay of 0.01. 

In the implementation related to GPT models, we utilized the API provided by OpenAI, configuring the temperature parameter to 0.7. All the codes for our experiments are public available via: \url{https://github.com/JiangMeng-JM/Cognitive-Pathways---Deep-Learning}.

\section{Results} \label{sec:results}

\subsection{Hierarchical text classification}
We report the model performance of the HTC task at both the parent and child nodes, as well as the overall performance by averaging them. The results can be seen in Table~\ref{tab:result_htc}. 
We also present the performance metrics of the ERNIE 3.0 model of each parent and child node as shown in Table~\ref{tab:results_htc_nodes}. The detailed performance of the GPT model family is not included due to their poor results.

From the experimental results, we find that each model performs better on the parent node than on the child nodes. ERNIE 3.0 significantly outperforms the GPT series models, suggesting that a supervised learning model is still necessary for the task of cognitive pathways extraction. From Table~\ref{tab:results_htc_nodes}, we can see that the ERNIE 3.0 model has an F1-score of more than 77\% on the parent node A, B, and C, and performs worse on the D node, with an F1 score of only 34.23\%. For supervised learning, the parent node A, B, and C has more data, which makes it easier for the model to learn, while the D node has less data, so it cannot learn adequately. For the child nodes, there is the problem of imbalance between classes, so the model may not be able to fully learn the features of fewer samples. As shown in Table~\ref{tab:dataset}, the categories ``All-or-nothing thinking'', ``Mental filter'' and ``Magnification and minimization'' under (B) Belief node have only about 50 sentences of training data, which results in these three categories not being able to make correct predictions.

\begin{table}
\centering
\caption{This table reports on the performance of the cognitive pathways extraction task (hierarchical text classification). The results are presented at both the parent and child node levels, as well as the average performance across all nodes (Overall). The metrics of precision, recall, and F1 score are reported as micro-averages.}
\label{tab:result_htc}
\scalebox{0.9}{
\begin{tabular}{c|l|clc} 
\hline
Models                   & \multicolumn{1}{c|}{Level}                  & Precision      & Recall         & \multicolumn{1}{l}{F1}     \\ 
\hline
\multirow{3}{*}{ERNIE 3.0}   & \multicolumn{1}{c|}{Parent nodes}           & \textbf{74.92} & \textbf{77.29} & \textbf{76.64}             \\
                         & Child nodes                                 & 40.21          & 41.33          & 46.70                      \\
                         & Overall                                     & 66.30          & 58.83          & \multicolumn{1}{l}{62.34}  \\ 
\hline
\multirow{3}{*}{GPT-4}   & \multicolumn{1}{c|}{Parent nodes}           & 31.69          & 41.47          & 35.93                      \\
                         & Child nodes                                 & 15.44          & 19.45          & 17.22                      \\
                         & \textcolor[rgb]{0.051,0.051,0.051}{Overall} & 23.43          & 30.07          & \multicolumn{1}{l}{26.34}  \\ 
\hline
\multirow{3}{*}{GPT-3.5} & \multicolumn{1}{c|}{Parent nodes}           & 28.61          & 36.50          & 32.08                      \\
                         & Child nodes                                 & 12.19          & 14.92          & 13.42                      \\
                         & Overall                                     & 20.27          & 25.31          & \multicolumn{1}{l}{22.51}  \\
\hline
\end{tabular}
}
\end{table}

\begin{table}
\centering
\caption{Performance of the ERNIE 3.0 model for the cognitive pathways hierarchical text classification task at the node level.}
\label{tab:results_htc_nodes}
\scalebox{0.9}{
\begin{tabular}{c|ccc} 
\hline
\textbf{\textbf{Nodes}}        & \multicolumn{1}{l}{Precision} & \multicolumn{1}{l}{Recall} & \multicolumn{1}{l}{F1}  \\ 
\hline
\textbf{(A) Activating Event}  & 80.58                         & 80.06                      & 80.32                   \\ 
\hline
Disease symptom                & 83.33                         & 72.46                      & 77.52                   \\
Social relation                & 58.46                         & 54.29                      & 56.30                   \\
Life                           & 55.77                         & 42.65                      & 48.33                   \\
Study and work                 & 73.02                         & 76.67                      & 74.80                   \\
Emotional                      & 58.14                         & 48.08                      & 52.63                   \\ 
\hline
\textbf{(B) Belief}            & 74.11                         & 82.25                      & 77.97                   \\ 
\hline
All-or-nothing thinking            & 0.00                          & 0.00                       & 0.00                    \\
Over-generalization             & 11.11                         & 5.71                       & 7.55                    \\
Mental filter               & 0.00                          & 0.00                       & 0.00                    \\
Disqualifying the positive         & 23.26                         & 25.64                      & 24.39                   \\
Jumping to~conclusions         & 30.19                         & 20.78                      & 24.62                   \\
Magnification and~minimization & 0.00                          & 0.00                       & 0.00                    \\
Emotional reasoning            & 21.43                         & 12.16                      & 15.52                   \\
Should statements              & 66.67                         & 80.00                      & 72.73                   \\
Labeling and mislabeling                       & 78.18                         & 67.19                      & 72.27                   \\
Blaming oneself/others         & 50.00                         & 16.67                      & 25.00                   \\ 
\hline
\textbf{(C) Consequence}      & 82.13                         & 76.58                      & 79.25                   \\ 
\hline
Emotional effect               & 56.10                         & 56.10                      & 56.10                   \\
Behavioral effect              & 62.82                         & 49.49                      & 55.37                   \\ 
\hline
\textbf{(D) Disputation}       & 39.58                         & 30.16                      & 34.23                   \\ 
\hline
Habitual disputation           & 18.52                         & 15.15                      & 16.67                   \\
Effective~disputation          & 36.36                         & 12.90                      & 19.05                   \\
\hline
\end{tabular}
}
\end{table}

\subsection{Text summarization}

The second task is to summarise the text of the extracted cognitive pathways, the detailed results are shown in Table~\ref{tab:result_text_sum}. We can find that GPT-based models outperform the fine-tuned supervised learning model Pegasus in this task. And the performance of GPT-3.5 and GPT-4 models are comparable.

GPT-4 achieved the highest scores on all evaluation metrics, indicating that it performs best in understanding the content of the original text and generating coherent and well-structured summaries.
GPT-3.5 scored relatively high on Rouge-L and BLEU-4, better than Pegasus but slightly lower than GPT-4. This means that GPT-3.5 does a good job of generating smoother and more accurate summaries, but is slightly less good at capturing detail and structure.
Pegasus had the lowest performance on all the metrics, nevertheless, it scored similar to GPTs on Rouge-2, suggesting that in some cases it was able to handle sentence structure and coherence better.
Overall, the GPTs performed optimally on the text summarisation task, demonstrating their strong capability in handling natural language comprehension and generation tasks. But because of its potentially hallucinatory output, the exact trade-off between the two is debatable.

\begin{table}
\centering
\caption{Result for text summarization task}
\label{tab:result_text_sum}
\scalebox{1}{
\begin{tabular}{c|cccc} 
\hline
Models           & \multicolumn{1}{l}{Rouge-1} & \multicolumn{1}{l}{Rouge-2} & \multicolumn{1}{r}{Rouge-L} & \multicolumn{1}{l}{BLEU-4}  \\ 
\hline
Pegasus & 47.49                       & 29.48                       & 39.33                       & 6.06                        \\
GPT-3.5          & 53.86                       & 30.11                       & 48.05                       & 8.62                        \\
GPT-4            & 54.92                       & 30.86                       & 48.65                       & 9.4                         \\
\hline
\end{tabular}
}
\end{table}

\begin{figure*}[!t]
\centering
\includegraphics[width=0.8\linewidth]{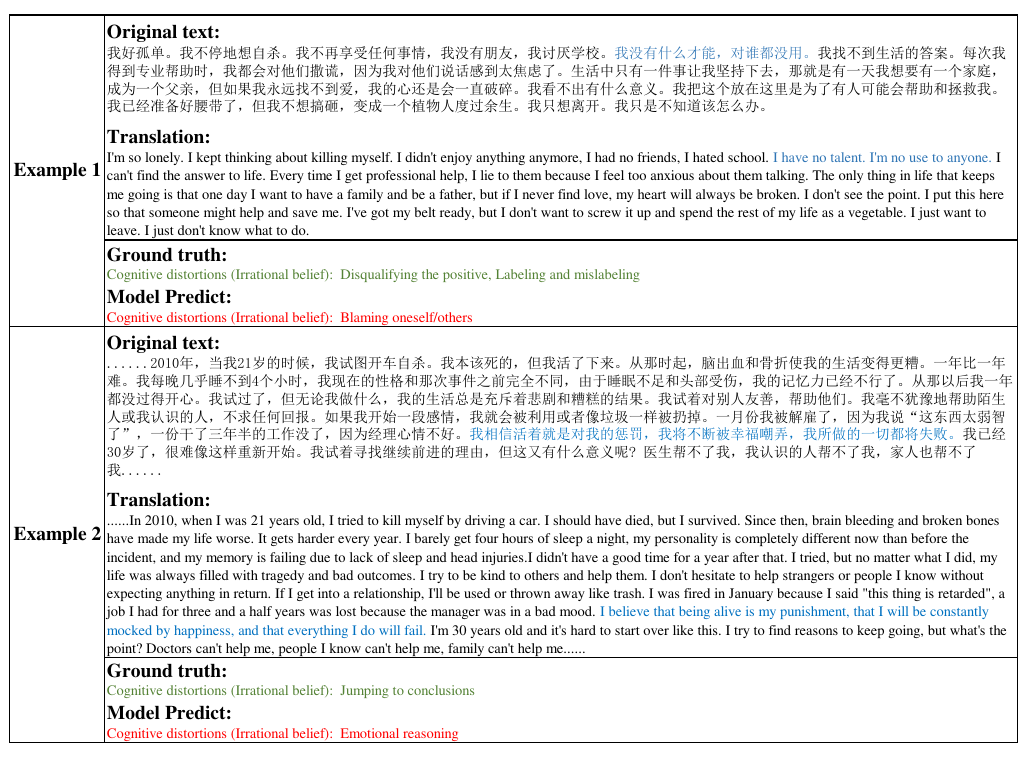}
\caption{Examples of errors in the hierarchical text classification task using the ERNIE 3.0 model are presented. For a clean presentation, we have only presented the portion relevant to the error and have not depicted the complete cognitive pathway extraction results.}
\label{fig:error_analysis_htc}
\end{figure*}

\begin{figure*}[!t]
\centering
\includegraphics[width=0.8\linewidth]{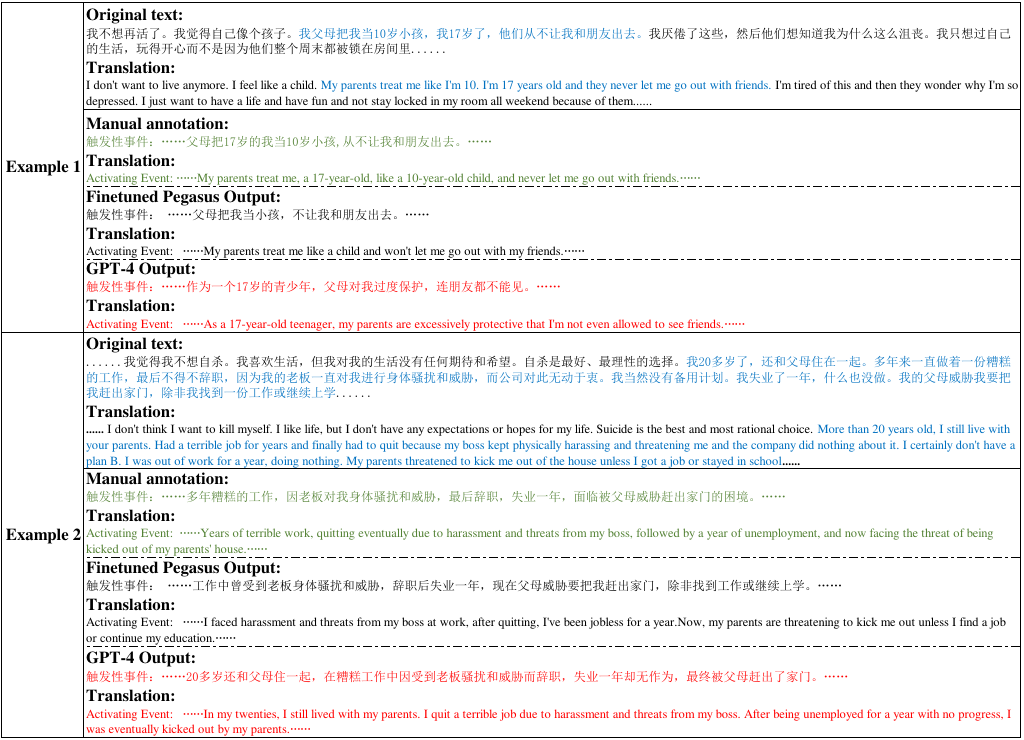}
\caption{Output of different models in the text summarization task. }
\label{fig:error_analysis_summary_llm}
\end{figure*}

\section{Discussion} \label{sec:discussion}

In the work on extracting cognitive pathways, the supervised learning model ERNIE 3.0 has shown superior performance to GPT-based models. It excels at analysing parent nodes, but encounters difficulties at the child nodes level due to insufficient training data. 
As illustrated in Example~1 in Figure~\ref{fig:error_analysis_htc}, sentences that should have been classified as ``Disqualifying the positive'' and ``Labeling and mislabeling'' are erroneously predicted as the category ``Blaming oneself/others.'' These errors often occur because while the parent nodes in the dataset have enough samples, the child nodes do not, resulting in insufficient training and prediction errors. 
Future research will explore the potential of multi-model hybrid decision strategies. This will involve developing specific models for parent node classifications, followed by collecting and using data to individually train models for child nodes within each parent category. Despite the expected increase in training costs, such an approach is expected to significantly improve performance.

For text summarisation task, GPT-based models have shown impressive results, outperforming the PEGASUS model on the Rouge-1, Rouge-L and BLEU-4 metrics. 
However, during the process of summary generation by LLMs, instances of input-conflicting hallucination arise. As depicted in Example~1 in Figure~\ref{fig:error_analysis_summary_llm}, GPT summarized that his parents wouldn't let him/her meet his/her friends because they were overprotective.
And in Figure~\ref{fig:error_analysis_summary_llm}~Example~2, The original sentence states, ``My parents threatened to kick me out of the house'', but the GPT summarizes it as ``Eventually my parents kicked me out of the house''.
In both cases, the GPT overly abstracts the content, leading to inaccurate results that could mislead the reader.
Although the summaries produced by the Pegasus model did not score as highly on evaluation metrics as those generated by the GPT, they do not exhibit problems with hallucinatory content.
Given the nature of the field of psychology, the question of tradeoffs between LLMs and deep learning technology choices is important.

\section{Conclusion} \label{sec:conclusion}

In this research, we explore the application of deep learning and LLMs techniques to identify cognitive pathways from social media content, guided by a cognitive psychology framework. This work aims to provide psychotherapists with a rapid means of reading the cognitive pathways in clients' statements, thereby enhancing the effectiveness of CBT. We categorize this challenge as a combination of hierarchical text classification and text summarization tasks. Our results show that while deep learning methods have a clear advantage in delineating cognitive pathways, LLMs show promising capabilities, particularly in text summarization. However, the problem of hallucination in LLMs cannot be ignored, and fake content may mislead the psychotherapists, so the choice of a specific solution needs to be decided on a case-by-case basis. Our study explores new applications of AI algorithms in CBT, and as the model is open source, it has the potential for a wide range of applications.

\FloatBarrier

\bibliography{references.bib}

\begin{thebibliography}{10}
\providecommand{\url}[1]{#1}
\csname url@samestyle\endcsname
\providecommand{\newblock}{\relax}
\providecommand{\bibinfo}[2]{#2}
\providecommand{\BIBentrySTDinterwordspacing}{\spaceskip=0pt\relax}
\providecommand{\BIBentryALTinterwordstretchfactor}{4}
\providecommand{\BIBentryALTinterwordspacing}{\spaceskip=\fontdimen2\font plus
\BIBentryALTinterwordstretchfactor\fontdimen3\font minus \fontdimen4\font\relax}
\providecommand{\BIBforeignlanguage}[2]{{%
\expandafter\ifx\csname l@#1\endcsname\relax
\typeout{** WARNING: IEEEtran.bst: No hyphenation pattern has been}%
\typeout{** loaded for the language `#1'. Using the pattern for}%
\typeout{** the default language instead.}%
\else
\language=\csname l@#1\endcsname
\fi
#2}}
\providecommand{\BIBdecl}{\relax}
\BIBdecl

\bibitem{WHO}
{World Health Organization}, ``Depressive disorder (depression),'' {World Health Organization}, 2023.

\bibitem{huang2019prevalence}
Y.~Huang, Y.~Wang, H.~Wang, Z.~Liu, X.~Yu, J.~Yan, Y.~Yu, C.~Kou, X.~Xu, J.~Lu \emph{et~al.}, ``Prevalence of mental disorders in china: a cross-sectional epidemiological study,'' \emph{The Lancet Psychiatry}, vol.~6, no.~3, pp. 211--224, 2019.

\bibitem{robinson2016social}
J.~Robinson, G.~Cox, E.~Bailey, S.~Hetrick, M.~Rodrigues, S.~Fisher, and H.~Herrman, ``Social media and suicide prevention: a systematic review,'' \emph{Early intervention in psychiatry}, vol.~10, no.~2, pp. 103--121, 2016.

\bibitem{cuijpers2023cognitive}
P.~Cuijpers, C.~Miguel, M.~Harrer, C.~Y. Plessen, M.~Ciharova, D.~Ebert, and E.~Karyotaki, ``Cognitive behavior therapy vs. control conditions, other psychotherapies, pharmacotherapies and combined treatment for depression: a comprehensive meta-analysis including 409 trials with 52,702 patients,'' \emph{World Psychiatry}, vol.~22, no.~1, pp. 105--115, 2023.

\bibitem{ellis2007practice}
A.~Ellis and W.~Dryden, \emph{The practice of rational emotive behavior therapy}.\hskip 1em plus 0.5em minus 0.4em\relax Springer publishing company, 2007.

\bibitem{stein2022psychiatric}
D.~J. Stein, S.~J. Shoptaw, D.~V. Vigo, C.~Lund, P.~Cuijpers, J.~Bantjes, N.~Sartorius, and M.~Maj, ``Psychiatric diagnosis and treatment in the 21st century: paradigm shifts versus incremental integration,'' \emph{World Psychiatry}, vol.~21, no.~3, pp. 393--414, 2022.

\bibitem{young2018recent}
T.~Young, D.~Hazarika, S.~Poria, and E.~Cambria, ``Recent trends in deep learning based natural language processing,'' \emph{ieee Computational intelligenCe magazine}, vol.~13, no.~3, pp. 55--75, 2018.

\bibitem{otter2020survey}
D.~W. Otter, J.~R. Medina, and J.~K. Kalita, ``A survey of the usages of deep learning for natural language processing,'' \emph{IEEE transactions on neural networks and learning systems}, vol.~32, no.~2, pp. 604--624, 2020.

\bibitem{alharbi2019twitter}
A.~S.~M. Alharbi and E.~de~Doncker, ``Twitter sentiment analysis with a deep neural network: An enhanced approach using user behavioral information,'' \emph{Cognitive Systems Research}, vol.~54, pp. 50--61, 2019.

\bibitem{nandwani2021review}
P.~Nandwani and R.~Verma, ``A review on sentiment analysis and emotion detection from text,'' \emph{Social network analysis and mining}, vol.~11, no.~1, p.~81, 2021.

\bibitem{tadesse2019detection}
M.~M. Tadesse, H.~Lin, B.~Xu, and L.~Yang, ``Detection of suicide ideation in social media forums using deep learning,'' \emph{Algorithms}, vol.~13, no.~1, p.~7, 2019.

\bibitem{fu2021distant}
G.~Fu, C.~Song, J.~Li, Y.~Ma, P.~Chen, R.~Wang, B.~X. Yang, and Z.~Huang, ``Distant supervision for mental health management in social media: suicide risk classification system development study,'' \emph{Journal of medical internet research}, vol.~23, no.~8, p. e26119, 2021.

\bibitem{singh2020realising}
D.~Singh and S.~Singh, ``Realising transfer learning through convolutional neural network and support vector machine for mental task classification,'' \emph{Electronics Letters}, vol.~56, no.~25, pp. 1375--1378, 2020.

\bibitem{aragon2023disorbert}
M.~Aragon, A.~P.~L. Monroy, L.~Gonzalez, D.~E. Losada, and M.~Montes, ``{DisorBERT: A Double Domain Adaptation Model for Detecting Signs of Mental Disorders in Social Media},'' in \emph{Proceedings of the 61st Annual Meeting of the Association for Computational Linguistics (Volume 1: Long Papers)}, 2023, pp. 15\,305--15\,318.

\bibitem{zhai2024chinese}
W.~Zhai, H.~Qi, Q.~Zhao, J.~Li, Z.~Wang, H.~Wang, B.~X. Yang, and G.~Fu, ``{Chinese MentalBERT: Domain-Adaptive Pre-training on Social Media for Chinese Mental Health Text Analysis},'' \emph{arXiv preprint arXiv:2402.09151}, 2024.

\bibitem{he2023towards}
T.~He, G.~Fu, Y.~Yu, F.~Wang, J.~Li, Q.~Zhao, C.~Song, H.~Qi, D.~Luo, H.~Zou \emph{et~al.}, ``Towards a psychological generalist ai: A survey of current applications of large language models and future prospects,'' \emph{arXiv preprint arXiv:2312.04578}, 2023.

\bibitem{qi2023evaluating}
H.~Qi, Q.~Zhao, C.~Song, W.~Zhai, D.~Luo, S.~Liu, Y.~J. Yu, F.~Wang, H.~Zou, B.~X. Yang \emph{et~al.}, ``Evaluating the efficacy of supervised learning vs large language models for identifying cognitive distortions and suicidal risks in {Chinese} social media,'' \emph{arXiv preprint arXiv:2309.03564}, 2023.

\bibitem{ernie3_sun2021ernie}
Y.~Sun, S.~Wang, S.~Feng, S.~Ding, C.~Pang, J.~Shang, J.~Liu, X.~Chen, Y.~Zhao, Y.~Lu \emph{et~al.}, ``{ERNIE} 3.0: Large-scale knowledge enhanced pre-training for language understanding and generation,'' \emph{arXiv preprint arXiv:2107.02137}, 2021.

\bibitem{pegasus_zhang2020pegasus}
J.~Zhang, Y.~Zhao, M.~Saleh, and P.~Liu, ``Pegasus: Pre-training with extracted gap-sentences for abstractive summarization,'' in \emph{International Conference on Machine Learning}.\hskip 1em plus 0.5em minus 0.4em\relax PMLR, 2020, pp. 11\,328--11\,339.

\bibitem{openai2023gpt4}
OpenAI, ``{Gpt-4 technical report},'' 2023.

\bibitem{beck1989cognitive}
A.~Beck and M.~Weishaar, ``Cognitive therapy. {Comprehensive} handbook of cognitive therapy,'' 1989.

\bibitem{hofmann2012efficacy}
S.~G. Hofmann, A.~Asnaani, I.~J. Vonk, A.~T. Sawyer, and A.~Fang, ``The efficacy of cognitive behavioral therapy: A review of meta-analyses,'' \emph{Cognitive therapy and research}, vol.~36, pp. 427--440, 2012.

\bibitem{tolin2010cognitive}
D.~F. Tolin, ``Is cognitive--behavioral therapy more effective than other therapies?: A meta-analytic review,'' \emph{Clinical psychology review}, vol.~30, no.~6, pp. 710--720, 2010.

\bibitem{sarracino2017rebt}
D.~Sarracino, G.~Dimaggio, R.~Ibrahim, R.~Popolo, S.~Sassaroli, and G.~M. Ruggiero, ``When {REBT} goes difficult: applying {ABC-DEF} to personality disorders,'' \emph{Journal of Rational-Emotive \& Cognitive-Behavior Therapy}, vol.~35, pp. 278--295, 2017.

\bibitem{selva2021albert}
J.~Selva, ``Albert {Ellis’} {ABC} model in the cognitive behavioral therapy spotlight,'' \emph{PositivePsychology. Retrieved from}, 2021.

\bibitem{burns1999feeling}
D.~D. Burns and A.~T. Beck, ``Feeling good: The new mood therapy,'' 1999.

\bibitem{payne2010increasing}
K.~A. Payne and G.~Myhr, ``Increasing access to cognitive-behavioural therapy ({CBT}) for the treatment of mental illness in {Canada}: a research framework and call for action,'' \emph{Healthcare Policy}, vol.~5, no.~3, p. e173, 2010.

\bibitem{foreman2016cognitive}
E.~I. Foreman and C.~Pollard, \emph{Cognitive Behavioural Therapy (CBT): Your Toolkit to Modify Mood, Overcome Obstructions and Improve Your Life}.\hskip 1em plus 0.5em minus 0.4em\relax Icon Books, Limited, 2016.

\bibitem{bailey2018theoretical}
E.~Bailey, S.~Rice, J.~Robinson, M.~Nedeljkovic, and M.~Alvarez-Jimenez, ``Theoretical and empirical foundations of a novel online social networking intervention for youth suicide prevention: A conceptual review,'' \emph{Journal of affective disorders}, vol. 238, pp. 499--505, 2018.

\bibitem{heng2021rewind}
Y.~K. Heng, ``{ReWIND}: psychoeducation game leveraging cognitive behavioral therapy ({CBT}) to enhance emotion control for generalized anxiety disorder,'' in \emph{Extended Abstracts of the 2021 CHI Conference on Human Factors in Computing Systems}, 2021, pp. 1--5.

\bibitem{liu2021efficacy}
Z.~Liu, D.~Qiao, Y.~Xu, W.~Zhao, Y.~Yang, D.~Wen, X.~Li, X.~Nie, Y.~Dong, S.~Tang \emph{et~al.}, ``The efficacy of computerized cognitive behavioral therapy for depressive and anxiety symptoms in patients with {COVID-19}: randomized controlled trial,'' \emph{Journal of medical Internet research}, vol.~23, no.~5, p. e26883, 2021.

\bibitem{brown2020language}
T.~Brown, B.~Mann, N.~Ryder, M.~Subbiah, J.~D. Kaplan, P.~Dhariwal, A.~Neelakantan, P.~Shyam, G.~Sastry, A.~Askell \emph{et~al.}, ``Language models are few-shot learners,'' \emph{Advances in neural information processing systems}, vol.~33, pp. 1877--1901, 2020.

\bibitem{graham2019artificial}
S.~Graham, C.~Depp, E.~E. Lee, C.~Nebeker, X.~Tu, H.-C. Kim, and D.~V. Jeste, ``Artificial intelligence for mental health and mental illnesses: an overview,'' \emph{Current psychiatry reports}, vol.~21, pp. 1--18, 2019.

\bibitem{vuyyuru2023transformer}
V.~A. Vuyyuru, G.~V. Krishna, S.~S.~C. Mary, S.~Kayalvili, and A.~M.~S. Alsubayhay, ``{A Transformer-CNN Hybrid Model for Cognitive Behavioral Therapy in Psychological Assessment and Intervention for Enhanced Diagnostic Accuracy and Treatment Efficiency},'' \emph{International Journal of Advanced Computer Science and Applications}, vol.~14, no.~7, 2023.

\bibitem{fitzpatrick2017delivering}
K.~K. Fitzpatrick, A.~Darcy, and M.~Vierhile, ``Delivering cognitive behavior therapy to young adults with symptoms of depression and anxiety using a fully automated conversational agent ({Woebot}): a randomized controlled trial,'' \emph{JMIR mental health}, vol.~4, no.~2, p. e7785, 2017.

\bibitem{rani2023mental}
K.~Rani, H.~Vishnoi, and M.~Mishra, ``{A Mental Health Chatbot Delivering Cognitive Behavior Therapy and Remote Health Monitoring Using NLP And AI},'' in \emph{2023 International Conference on Disruptive Technologies (ICDT)}.\hskip 1em plus 0.5em minus 0.4em\relax IEEE, 2023, pp. 313--317.

\bibitem{fu2023enhancing}
G.~Fu, Q.~Zhao, J.~Li, D.~Luo, C.~Song, W.~Zhai, S.~Liu, F.~Wang, Y.~Wang, L.~Cheng \emph{et~al.}, ``Enhancing psychological counseling with large language model: A multifaceted decision-support system for non-professionals,'' \emph{arXiv preprint arXiv:2308.15192}, 2023.

\bibitem{sharma2023cognitive}
A.~Sharma, K.~Rushton, I.~Lin, D.~Wadden, K.~Lucas, A.~Miner, T.~Nguyen, and T.~Althoff, ``Cognitive reframing of negative thoughts through human-language model interaction,'' in \emph{Proceedings of the 61st Annual Meeting of the Association for Computational Linguistics (Volume 1: Long Papers)}, 2023, pp. 9977--10\,000.

\bibitem{sharma2023facilitating}
A.~Sharma, K.~Rushton, I.~W. Lin, T.~Nguyen, and T.~Althoff, ``Facilitating self-guided mental health interventions through human-language model interaction: A case study of cognitive restructuring,'' \emph{arXiv preprint arXiv:2310.15461}, 2023.

\bibitem{huibers2021road}
M.~J. Huibers, L.~Lorenzo-Luaces, P.~Cuijpers, and N.~Kazantzis, ``On the road to personalized psychotherapy: A research agenda based on cognitive behavior therapy for depression,'' \emph{Frontiers in Psychiatry}, vol.~11, p. 607508, 2021.

\bibitem{minaee2021deep}
S.~Minaee, N.~Kalchbrenner, E.~Cambria, N.~Nikzad, M.~Chenaghlu, and J.~Gao, ``Deep learning--based text classification: a comprehensive review,'' \emph{ACM computing surveys (CSUR)}, vol.~54, no.~3, pp. 1--40, 2021.

\bibitem{silla2011survey}
C.~N. Silla and A.~A. Freitas, ``A survey of hierarchical classification across different application domains,'' \emph{Data mining and knowledge discovery}, vol.~22, pp. 31--72, 2011.

\bibitem{zhou2020hierarchy}
J.~Zhou, C.~Ma, D.~Long, G.~Xu, N.~Ding, H.~Zhang, P.~Xie, and G.~Liu, ``Hierarchy-aware global model for hierarchical text classification,'' in \emph{Proceedings of the 58th annual meeting of the association for computational linguistics}, 2020, pp. 1106--1117.

\bibitem{peng2018large}
H.~Peng, J.~Li, Y.~He, Y.~Liu, M.~Bao, L.~Wang, Y.~Song, and Q.~Yang, ``Large-scale hierarchical text classification with recursively regularized deep graph-cnn,'' in \emph{Proceedings of the 2018 world wide web conference}, 2018, pp. 1063--1072.

\bibitem{mao2019hierarchical}
Y.~Mao, J.~Tian, J.~Han, and X.~Ren, ``{Hierarchical Text Classification with Reinforced Label Assignment},'' in \emph{Proceedings of the 2019 Conference on Empirical Methods in Natural Language Processing and the 9th International Joint Conference on Natural Language Processing (EMNLP-IJCNLP)}, 2019, pp. 445--455.

\bibitem{wehrmann2018hierarchical}
J.~Wehrmann, R.~Cerri, and R.~Barros, ``Hierarchical multi-label classification networks,'' in \emph{International conference on machine learning}.\hskip 1em plus 0.5em minus 0.4em\relax PMLR, 2018, pp. 5075--5084.

\bibitem{meng2019weakly}
Y.~Meng, J.~Shen, C.~Zhang, and J.~Han, ``Weakly-supervised hierarchical text classification,'' in \emph{Proceedings of the AAAI conference on artificial intelligence}, vol.~33, no.~01, 2019, pp. 6826--6833.

\bibitem{han2021pre}
X.~Han, Z.~Zhang, N.~Ding, Y.~Gu, X.~Liu, Y.~Huo, J.~Qiu, Y.~Yao, A.~Zhang, L.~Zhang \emph{et~al.}, ``Pre-trained models: Past, present and future,'' \emph{AI Open}, vol.~2, pp. 225--250, 2021.

\bibitem{chiorrini2021emotion}
A.~Chiorrini, C.~Diamantini, A.~Mircoli, and D.~Potena, ``Emotion and sentiment analysis of tweets using {BERT}.'' in \emph{EDBT/ICDT Workshops}, vol.~3, 2021.

\bibitem{chopra2016abstractive}
S.~Chopra, M.~Auli, and A.~M. Rush, ``Abstractive sentence summarization with attentive recurrent neural networks,'' in \emph{Proceedings of the 2016 conference of the North American chapter of the association for computational linguistics: human language technologies}, 2016, pp. 93--98.

\bibitem{rush2015neural}
\BIBentryALTinterwordspacing
A.~M. Rush, S.~Chopra, and J.~Weston, ``A neural attention model for abstractive sentence summarization,'' in \emph{Proceedings of the 2015 Conference on Empirical Methods in Natural Language Processing}, L.~M{\`a}rquez, C.~Callison-Burch, and J.~Su, Eds.\hskip 1em plus 0.5em minus 0.4em\relax Lisbon, Portugal: Association for Computational Linguistics, Sep. 2015, pp. 379--389. [Online]. Available: \url{https://aclanthology.org/D15-1044}
\BIBentrySTDinterwordspacing

\bibitem{transformer_vaswani2017attention}
A.~Vaswani, N.~Shazeer, N.~Parmar, J.~Uszkoreit, L.~Jones, A.~N. Gomez, {\L}.~Kaiser, and I.~Polosukhin, ``Attention is all you need,'' \emph{Advances in neural information processing systems}, vol.~30, 2017.

\bibitem{radford2018improving}
A.~Radford, K.~Narasimhan, T.~Salimans, I.~Sutskever \emph{et~al.}, ``Improving language understanding by generative pre-training,'' 2018.

\bibitem{raffel2020exploring}
C.~Raffel, N.~Shazeer, A.~Roberts, K.~Lee, S.~Narang, M.~Matena, Y.~Zhou, W.~Li, and P.~J. Liu, ``Exploring the limits of transfer learning with a unified text-to-text transformer,'' \emph{The Journal of Machine Learning Research}, vol.~21, no.~1, pp. 5485--5551, 2020.

\bibitem{lewis2020bart}
M.~Lewis, Y.~Liu, N.~Goyal, M.~Ghazvininejad, A.~Mohamed, O.~Levy, V.~Stoyanov, and L.~Zettlemoyer, ``{BART}: Denoising sequence-to-sequence pre-training for natural language generation, translation, and comprehension,'' in \emph{Proceedings of the 58th Annual Meeting of the Association for Computational Linguistics}, 2020, pp. 7871--7880.

\bibitem{goyal2022news}
T.~Goyal, J.~J. Li, and G.~Durrett, ``News summarization and evaluation in the era of gpt-3,'' \emph{arXiv preprint arXiv:2209.12356}, 2022.

\bibitem{Pu2023summarization}
X.~Pu, M.~Gao, and X.~Wan, ``Summarization is (almost) dead,'' \emph{arXiv preprint arXiv:2309.09558}, 2023.

\bibitem{xu2024hallucination}
Z.~Xu, S.~Jain, and M.~Kankanhalli, ``Hallucination is inevitable: An innate limitation of large language models,'' \emph{arXiv preprint arXiv:2401.11817}, 2024.

\bibitem{tonmoy2024comprehensive}
S.~Tonmoy, S.~Zaman, V.~Jain, A.~Rani, V.~Rawte, A.~Chadha, and A.~Das, ``A comprehensive survey of hallucination mitigation techniques in large language models,'' \emph{arXiv preprint arXiv:2401.01313}, 2024.

\bibitem{dai2019transformer}
Z.~Dai, Z.~Yang, Y.~Yang, J.~G. Carbonell, Q.~Le, and R.~Salakhutdinov, ``Transformer-{XL}: Attentive language models beyond a fixed-length context,'' in \emph{Proceedings of the 57th Annual Meeting of the Association for Computational Linguistics}, 2019, pp. 2978--2988.

\end{thebibliography}
\bibliographystyle{IEEEtran}
\balance

\end{document}